\documentclass{ecai} 

\usepackage{latexsym}
\usepackage{amssymb}
\usepackage{amsmath}
\usepackage{amsthm}
\usepackage{booktabs}
\usepackage{enumitem}
\usepackage{graphicx}
\usepackage{color}

\usepackage{amsmath,amssymb,amsfonts}
\usepackage{graphicx}
\usepackage{pgfplots}
\usepackage{stfloats}
\usepackage{subfig}
\usepackage{textcomp}
\usepackage{xcolor}
\usepackage{hyperref}
\usepackage[T1]{fontenc} 
\usepackage[utf8]{inputenc}
\usepackage{verbatim}
\usepackage{float}
\usepackage{multirow}
\usepackage{booktabs}
\usepackage{tikz}

\newcommand{\BibTeX}{B\kern-.05em{\sc i\kern-.025em b}\kern-.08em\TeX}

\newcommand{\mathescape}[1]{\relax\ifmmode#1\else $#1$\fi} 
\newcommand{\pnrank}[0]{\mathescape{\mathrm{PNRank}}}
\newcommand{\pnmax}[0]{\mathescape{\mathrm{PNMax}}}
\newcommand{\pnsum}[0]{\mathescape{\mathrm{PNSum}}}

\newcommand{\cpn}[0]{\mathescape{C_{pn}}}

\usepackage{algorithm,algorithmicx}
\usepackage[noend]{algpseudocode}
\MakeRobust{\Call}
 \newcommand{\LineIf}[2]{\State \algorithmicif\ {#1}\ \algorithmicthen\ {#2}}

\newcommand{\LineIfElseElse}[5]{\State \algorithmicif\ {#1}\ \algorithmicthen\ {#2} \State \algorithmicelse\ \algorithmicif\ {#3}\ \algorithmicthen\ {#4} \State \algorithmicelse\ {#5} }
\newcommand{\LineForAll}[2]{\State \algorithmicforall\ {#1}\ \algorithmicdo\ {#2}}
\newcommand{\DLineForAll}[2]{\State \algorithmicforall\ {#1}\ \algorithmicdo\ \State \hspace{\algorithmicindent} {#2}}

\begin{document}


\begin{frontmatter}


\paperid{9404} 


\title{Generalized Proof-Number Monte-Carlo Tree Search}


\author[A]{\fnms{Jakub}~\snm{Kowalski}\orcid{0000-0003-1932-4278}\thanks{Corresponding Author. Email: jakub.kowalski@cs.uni.wroc.pl}}
\author[B]{\fnms{Dennis J. N. J.}~\snm{Soemers}\orcid{0000-0003-3241-8957}}
\author[A]{\fnms{Szymon}~\snm{Kosakowski}\orcid{0009-0006-7992-0310}} 
\author[B]{\fnms{Mark H. M.}~\snm{Winands}\orcid{0000-0002-0125-0824}} 

\address[A]{University of Wroc{\l}aw}
\address[B]{Maastricht University}



\begin{abstract}
This paper presents Generalized Proof-Number Monte-Carlo Tree Search: a generalization of recently proposed combinations of Proof-Number Search (PNS) with Monte-Carlo Tree Search (MCTS), which use (dis)proof numbers to bias UCB1-based Selection strategies towards parts of the search that are expected to be easily (dis)proven. We propose three core modifications of prior combinations of PNS with MCTS. First, we track proof numbers per player. This reduces code complexity in the sense that we no longer need disproof numbers, and generalizes the technique to be applicable to games with more than two players. Second, we propose and extensively evaluate different methods of using proof numbers to bias the selection strategy, achieving strong performance with strategies that are simpler to implement and compute. Third, we merge our technique with Score Bounded MCTS, enabling the algorithm to prove and leverage upper and lower bounds on scores---as opposed to only proving wins or not-wins. Experiments demonstrate substantial performance increases, reaching the range of 80\% for 8 out of the 11 tested board games.



\end{abstract}

\end{frontmatter}

\section{Introduction}

Monte-Carlo Tree Search (MCTS)~\cite{coulom06,Kocsis_2006_Bandit} is a best-first search method guided by the results of Monte-Carlo simulations, well established in game AI \cite{mctssurvey,swiechowski2023monte}. Using the results of previous simulations, the method gradually builds up a game tree in memory and increasingly becomes better at accurately estimating the values of the most promising moves. MCTS has substantially advanced the state of the art in several deterministic game domains, in particular Go \cite{Silver2017mastering}, but also other board games including  Amazons~\cite{Lorentz08}, Hex \cite{arneson10}, Lines of Action \cite{winands10},  and   General Game Playing (GGP) \cite{bjornsson09}. 

In tactical games, where the main line towards the winning position is typically narrow with many non-progressing alternatives, MCTS may often lead to an erroneous outcome because the nodes' values in the tree do not converge fast enough to their game-theoretic value. To mitigate this effect, MCTS variants have been proposed that integrate minimax search \cite{winands08b,winands11,LanctotWPS14,baier2015}. Recently,  Proof-Number Search (PNS) \cite{allis94} has been integrated in MCTS \cite{Doe_2022_Combining,Randall2024}. PNS has the advantage proving endgames faster than traditional minimax in many domains. The variant PN-MCTS\cite{Kowalski2024ProofNumber} has been shown to improve over default MCTS in domains such as Lines of Action, MiniShogi, Knightthrough, and Awari.


In this paper, we propose a generalization of PN-MCTS, called Generalized Proof-Number Monte-Carlo Tree Search (GPN-MCTS).  First, this extension of PN-MCTS tracks proof numbers per player, which reduces code complexity in the sense that we no longer need disproof numbers, and generalizes the technique to be applicable to games with more than two players. Second, GPN-MCTS contains different enhancements of using proof numbers to bias the UCT selection strategy \cite{Kocsis_2006_Bandit}, achieving strong performance with strategies that are simpler to implement and compute. Third, PN-MCTS is integrated with Score Bounded MCTS \cite{cazenave2011ScoreBoundedMCTS}, enabling the technique to prove and leverage upper and lower bounds on scores---as opposed to only proving wins or non-wins.

For the purpose of the experiments, GPN-MCTS was implemented in the Ludii general game playing system \cite{Piette_2020_Ludii}.\footnote{After the reviewing process a pull request would be proposed to include GPN-MCTS into the official public repository of the Ludii project.} Although it is not guaranteed to work for every game, in many cases when it works, its improvements are significant, reaching 80\% win rate on 8 out of the 11 tested board games, and over 60\% on the remaining three. 
Moreover, we show that it pairs well with the Score-bounded MCTS enhancement.



This paper is organized as follows. First, Section \ref{sec:back} describes the foundational methodology for this paper, MCTS and PNS. Next,  Generalized Proof-Number MCTS is introduced in Section \ref{sec:GPM}. Subsequently, the approach is tested on a diverse set of games in Section \ref{sec:Exp}. Finally, the paper concludes and gives an outlook on future research in Section \ref{sec:conclusion}.



\section{Background}\label{sec:back}


\subsection{Monte Carlo Tree Search}

\begin{figure}[t]
    \centering
    \includegraphics[width=\columnwidth]{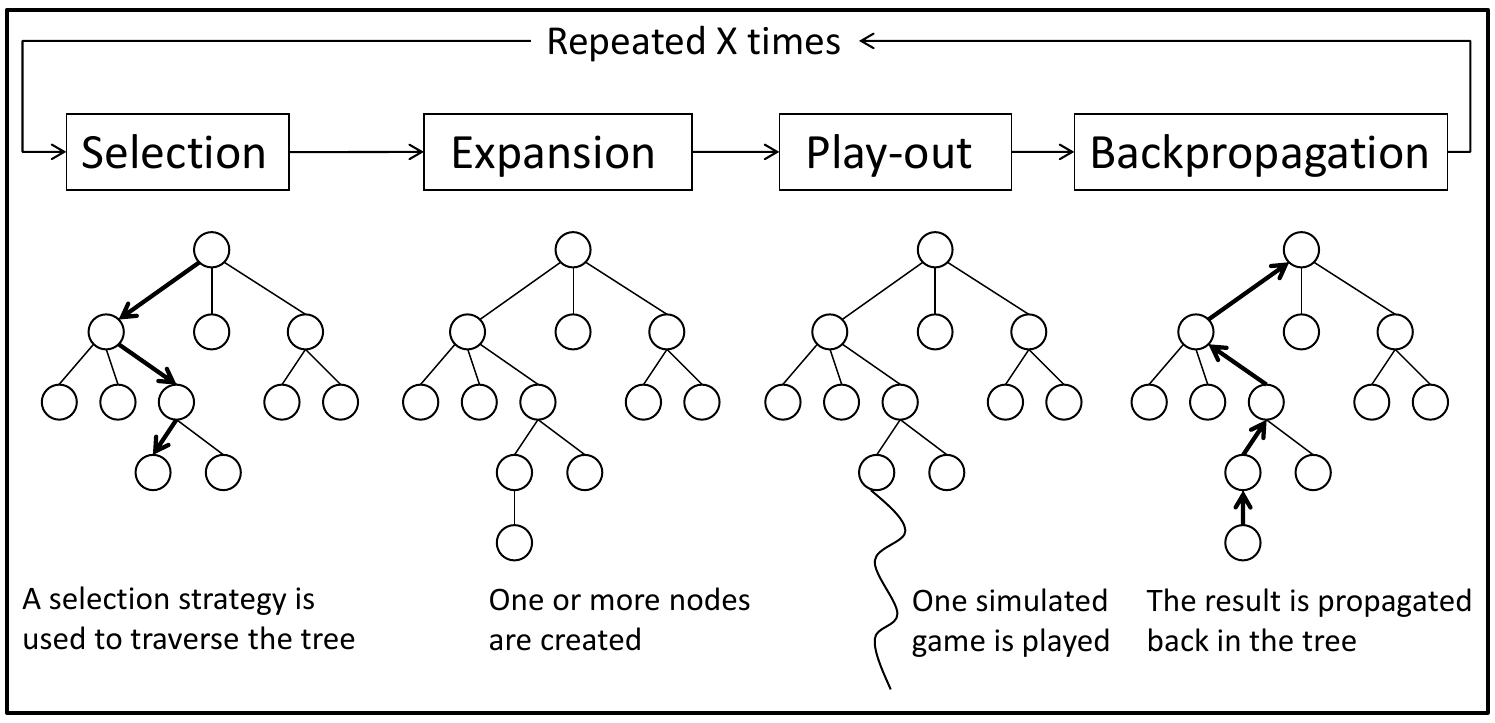}
    \caption{Outline of Monte-Carlo Tree Search.}
    \label{fig:mcts}
    \vspace{20pt}
\end{figure}

Monte Carlo Tree Search (MCTS) \cite{coulom06,Kocsis_2006_Bandit} is a best-first search method that gradually builds up a search tree, balancing exploitation of parts that seem promising based on earlier iterations, with explorations of parts that were infrequently explored. It does this by iterating through four strategic steps \cite{chaslot08}, depicted in \figurename~\ref{fig:mcts}, until a time or iteration budget expires.

\paragraph{Selection Step.} The selection step traverses the tree, starting from the root node, until a node is reached for which there are still legal actions that have not yet been expanded into nodes of the search tree (or until a terminal node is reached). This step implements the trade-off between exploration and exploitation. One of the most common baseline implementations of MCTS---referred to as \textit{Upper confidence Bounds applied to Trees} (UCT) \cite{Kocsis_2006_Bandit}---uses the UCB1 strategy \cite{auerfinit02} to choose among the children of any given current node. It works as follows. Let $I$ be the set of nodes immediately reachable from the
current node $p$. The selection strategy selects the child $b$ of
node $p$ that satisfies Formula~(\ref{eq:ucb1}):
\begin{equation}
\label{eq:ucb1}
\mathit{b\in \mathrm{argmax}_{i \in I} \left(v_i + C \times \sqrt{\frac{\ln{n_p}}{n_i}}\right)},
\end{equation}
where $v_i$ is the estimated value of the node $i$, $n_i$ is the visit count of $i$, and $n_p$ is the visit count of $p$. $C$ is a
hyperparameter which can be tuned experimentally. Here, ties are broken randomly.

\paragraph{Expansion Step.} As previously stated, the selection step continues until a node is reached that has not yet expanded all of its children. Among the children that have not been stored in the tree, one is selected uniformly at random. This node $L$ is then added as a new leaf node. If the selection step arrives at a terminal node, the expansion and subsequent play-out steps are skipped.

\paragraph{Play-out Step.} From the new leaf node $L$ onwards, the play-out step is performed. Moves are selected in self-play until the end of the game is reached. This step might consist of playing uniformly random moves or---often better---semi-random moves chosen according to a \textit{simulation strategy}.

\paragraph{Backpropagation Step.} In the final step, the result \textit{R}
of a play-out $k$ is backpropagated from the leaf node $L$, through the
previously traversed nodes, all the way up to the root. 
The result is scored positively $(R_k=+1)$ if the game is won, and negatively
$(R_k=-1)$ if the game is lost. Draws lead to a result $R_k=0$. A \emph{backpropagation strategy} is applied to the  \textit{value} $v_i$ of a node $i$. Here, it is computed by taking the average of the results of all simulated games made through this node \cite{coulom06}, i.e., $v_i=(\sum_{k \in K} R_k ) / n_i$, where $K$ is the set of indices for all play-outs. Visit counts $n_i$ for all nodes along the trajectory are also incremented.

When the search budget expires, the move that is ultimately selected to be played is the one from the root node that has the highest visit count (though other strategies are possible as well \cite{chaslot08}.

\subsection{Proof-Number Search}\label{sec:pns}

 Proof-Number Search (PNS) is a best-first search method especially suited for finding the
game-theoretic value in game trees \cite{allis94}. Its aim is to prove
a particular goal. In the context of this paper, the goal is to prove a forced win for the player to move. A tree can have three values:
\textit{true}, \textit{false}, or \textit{unknown}.  In case of a
forced win, the tree is \textit{proven}  and its value is true. In
 case of a forced loss or draw, the tree is \textit{disproven}
and its value is false. Otherwise, the value of the tree is unknown.
As long as the value of the root is unknown, the most-promising node
is expanded. Like MCTS, PNS
does not need a domain-dependent heuristic evaluation function to
determine the most-promising node \cite{allis94}. In PNS,  this
node is usually called the \textit{most-proving} node. PNS
selects the most-proving node using two criteria: (1) the shape of
the search tree (the branching factor of every internal node) and
(2) the values of the leaves. These two criteria enable PNS to
treat game trees with a non-uniform branching factor efficiently.

\begin{figure}[t]
\centerline{\includegraphics[width=\columnwidth]{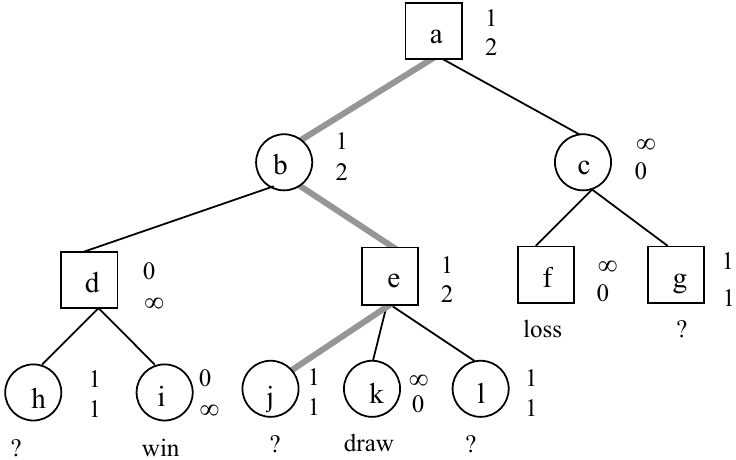} }
\caption{An AND/OR tree with proof and disproof numbers. A square denotes an OR node, and a circle denotes an AND node. The numbers to the right of a node
denote the proof number (upper) and disproof number  (lower).}
 \label{pntree}
 \vspace{20pt}
\end{figure}

PN search represents the game as an AND/OR tree.  OR nodes correspond to
positions with the first player to play, while in AND nodes, the second player is to play. An example of such a tree is given in Fig. \ref{pntree}.  In PNS, the \textit{proof number}
(\emph{pn}) represents the minimum number of leaf nodes, which have
to be proven in order to prove the node. Analogously, a
\textit{disproof number} (\emph{dpn}) represents the minimum number
of leaf nodes that have to be disproven in order to disprove the
node. Because the goal of the search is to prove a forced win, winning
nodes are regarded as proven. Therefore, they have $pn=0$ and
$dpn=\infty$. Lost or drawn
nodes are regarded disproven. They have $pn =\infty$ and $dpn=0$.
Unknown leaf nodes have  $pn=1$ and $dpn=1$.  The \emph{pn} of an
internal OR node is equal to the minimum of its children's proof
numbers, because to prove an OR node it suffices to prove one child.
The \emph{dpn} of an internal OR node is equal to the sum of
its children's disproof numbers, because to disprove an OR node all
the children have to be disproven. The  \emph{pn} of an internal AND node is equal to the sum of its children's
proof numbers, because to prove an AND node all the children have to
be proven. The \emph{dpn} of an AND node is equal to the
minimum of its children's disproof numbers, because to disprove an AND
node it suffices to disprove one child. 

The procedure of selecting the most-proving node to expand next is as
follows. The algorithm starts at the root. Then, at each OR node the child with
the smallest \emph{pn} is selected as successor, and at each AND
node the child with the smallest  \emph{dpn} is selected as
successor. Finally, when a leaf node is reached, it is expanded
(which makes the leaf node an internal node) and the newborn
children are evaluated.

\subsection{Proof-Number Monte-Carlo Tree Search}

Proof-Number Monte-Carlo Tree Search (PN-MCTS) was initially proposed in \cite{Doe_2022_Combining}, introducing an enhancement of the UCB1 formula for two-player zero-sum games that includes a PNS-related term biasing selection towards the children preferred by Proof-Number Search.
It was later extended by \cite{Kowalski2024ProofNumber} to take advantage of the observation that already computed (dis)proof numbers can also serve to bias final move selection and skip solved subtrees, similarly to a Score-Bounded MCTS \cite{cazenave2011ScoreBoundedMCTS}.
In this work, we aim to generalize on all the above accomplishments.

PN-MCTS tracks proof and disproof numbers from the point of view of the root node player in all nodes of the MCTS tree. The OR/AND nodes are assigned accordingly, from the root player's perspective.
The PNS-based part of the algorithm is a paranoid type \cite{Saito10} as opponent decisions, based on disproof numbers, make it more interested in preventing the root player's victory than maximizing its own result.

The proposed UCT-PN formula, shown in (\ref{eq:uctpn}), extends the standard UCB1 (\ref{eq:ucb1}), introducing a term that is greater for child nodes that are close rto being solved according to (dis)proof numbers, weighted by a \cpn\ constant.

 \begin{equation}
 \label{eq:uctpn}
 \mathit{b\in \mathrm{argmax}_{i \in \mathcal{I}} \left(v_i + C \times \sqrt{\frac{\ln{n_p}}{n_i}} + \cpn\times\pnrank(i, \mathcal{I}) \right)}
 \end{equation}

Instead of directly using the (dis)proof numbers in the formula, the idea is to sort their values and use a ranking system. This process aims to reflect the observation that the magnitudes of differences amongst (dis)proof numbers technically do not have much meaning, and is encapsulated by the \pnrank\ function (\ref{eq:pnrank}).

\begin{equation} \label{eq:pnrank}
 \pnrank(i, \mathcal{I})=  1 - \frac{\mathrm{rank}(i)}{\mathrm{max}_{j \in \mathcal{I}}\ \mathrm{rank}(j)}
\end{equation}

The best node---the one that would be chosen by PNS---gets a rank of 1. This is the child with the lowest proof number in OR nodes, and the one with the lowest disproof number in AND nodes.
The next one in order would get rank 2, and so on. Tied nodes are awarded the same rank. 
Then, the ranks are normalized into the range of $[0, 1]$, to allow easier scaling with the exploration and exploitation terms of the UCB1 formula.


Experiments performed in \cite{Kowalski2024ProofNumber} showed that the introduction of UCT-PN was enough to achieve an overwhelming performance (around 87\% versus the vanilla MCTS) in Lines of Action.
Further extension by final move selection and subtree solving were necessary to obtain winrates above 65\% for MiniShogi and Knighttrough.
To obtain a winning winrate for the last of the tested games, Awari, a special approach handling draws was introduced, involving the maintenance of a second set of (dis)proof numbers.




\section{Generalized Proof-Number MCTS}\label{sec:GPM}

In this section, we present the GPN-MCTS algorithm and discuss its main features.

\subsection{Proof Numbers per Player}

The goal of the PN-UCB formula is to bias exploration towards subtrees that can be proven quickly. Thus, a reasonable assumption is that each player is interested in their own win. 
As mentioned before, this is not the approach taken by the PN-MCTS version described in \cite{Kowalski2024ProofNumber}.
In that work, all (dis)proof numbers are computed for the perspective of the root player, with proof numbers being used to guide selection in OR nodes, and disproof numbers in AND nodes.
This approach results in asymmetry in behavior in particular in games where draws can occur, as the root player attempts to prove their wins, whereas the opponent only attempts to disprove the root player's wins (i.e., attempts to prove any mix of drawing and winning for themselves).

We propose a different approach, which consists of tracking separate proof numbers \emph{per player} in each node of the search tree.
This reduces code complexity in the sense that we no longer need to encode separate logic for handling disproof numbers, 
and the search behaviour becomes more symmetric in that each player attempts to prove their own wins.

This modification influences how the algorithm perceives AND/OR proof nodes.
Let $i$ be a game tree node, and $p$ the player to move in that node. 
To prove a win, it is enough that any of the available paths leads to a win, so it is an OR node \emph{for the player $p$}.
From the perspective of any players other than $p$, the same node is treated as an AND node, in which all children must be proven for the node to be considered proven.


The advantage of the approach with a single proof number per player is that it naturally generalizes for games with more than two players.
Although the proof number trees are no longer composed of alternating AND/OR layers, the way the algorithm behaves remains unchanged.
For each player, when computing their proof tree, the first-player player's nodes are OR nodes, and all the remaining ones are AND nodes.

That way, the algorithm still assumes that every player is interested in proving their own win.
However, proving a win for a given player still requires a paranoid assumption about the strategies of their opponents.
Thus, our algorithm in this part behaves as Paranoid Proof-Number Search \cite{Saito10}, a generalization of Proof-Number Search allowing to prove games with any number of players.

\subsection{UCT-PN Formulas}\label{sec:pnformulas}

The purpose of PN-term in the UCT-PN formula is to bias search towards nodes that are potentially easier to prove. 
However, the actual strength of this bias and how it depends on the actual proof numbers, is a critical point to discuss.
Formula (\ref{eq:pnrank}), using rank ordering of values, was proposed by \cite{Doe_2022_Combining,Kowalski2024ProofNumber} with a justification that the magnitudes of differences amongst (dis)proof numbers are not meaningful. 


We suspect that throwing away information about the actual (dis)proof values is a potential waste, and decided to search for possible alternatives. 
In fact, we believe that there are many functions that could serve this role, which may be simpler to implement and computationally cheaper than \pnrank.
This may be of special importance, as the problem of the \pnrank\ formula is its inefficiency. The need to sort all children's values every time a (dis)proof number is updated to calculate proper rank is a significant computational effort. 
Note that this default behavior may be optimized: either by taking advantage of partial ordering and only fixing the position of the updated element (in $O(|\mathcal{I}|)$ or by using a method described in Section~\ref{sec:backpropoptim}. 

In this paper, we introduce two alternative formulas that can be used instead: $\pnmax$ and $\pnsum$. We think both represent natural approaches when trying to take into account the actual proof values and making the spread proportionally rather than evenly.

\subsubsection{Expected properties}

We begin by establishing the desired properties of a proof-number-based bias function, so it will nicely fit the framework, and its type could be treated as a parameter of the GPN-MCTS.

Thus, in line with the structure of (\ref{eq:pnrank}), the proper formula should be a function taking a node $i$, and a set of nodes $\mathcal{I}$ (being $i$ and all its siblings) and returning a number in $[0, 1]$, with larger values representing a greater bias towards selecting the node. 

Additionally, we define a set of conditions that we argue a proper bias formula should meet.
Let $\mathrm{pn}(i)$ be the proof number of a node~$i$.

\begin{enumerate}[label=(\alph*)]
\item Formula should return 0 if it is impossible to prove a node ($\mathrm{pn}(i)$ is $\infty$).
\item If for any $j\in\mathcal{I}$, $\mathrm{pn}(j)$ is infinite, the values for all finite children should be strictly positive.
\item If $\mathrm{pn}(i)$ equals $\mathrm{pn}(j)$, then the output for $i$ and $j$, given the same $\mathcal{I}$ should also be equal.
\item (optional) Formula should always return 1 for the lowest finite $\mathrm{pn}(i)$ value among $\mathcal{I}$.
\end{enumerate}

The invariants above ensure that the actual behavior of the bias formula aligns with the natural expectations.
For the \pnrank\ formula defined as (\ref{eq:pnrank}), the first three conditions hold. 

\subsubsection{\pnmax}

The first of the newly proposed approaches is to scale values with respect to the range of (finite) proof values among the children. 
This bias formula, named $\pnmax$ (\ref{eq:pnmax}), requires calculating the minimal and maximal proof numbers among all children, which is significantly less computationally expensive than sorting.

The functions $\mathrm{maxf}$ and $\mathrm{minf}$ compute, respectively, the maximum and minimum proof values among the given set of nodes, taking into account only finite values. Thus, $\mathrm{maxf}(\mathcal{I}) = \mathrm{max}_{j \in \mathcal{I} \wedge \mathrm{pn}(j)\neq\infty} \mathrm{pn}(j)$.
\begin{equation} \label{eq:pnmax}
  \pnmax(i, \mathcal{I}) = \begin{cases} 
      0 & \mathrm{if}\  \mathrm{pn}(i)\ \mathrm{is}\ \infty \\
       1 - \frac{\mathrm{pn}(i)-\mathrm{minf}(\mathcal{I})}{1+\mathrm{maxf}(\mathcal{I})-\mathrm{minf}(\mathcal{I})}   & \mathrm{otherwise} \\
   \end{cases}
\end{equation}
For this formula, all four conditions are satisfied.

\subsubsection{\pnsum}
As an alternative, we propose a bias formula that spreads values proportionally with respect to the sum of all finite proof values among the children. For this formula, $\pnsum$ (\ref{eq:pnsum}), all conditions except (d) are met.

\begin{equation} \label{eq:pnsum}
\pnsum(i, \mathcal{I}) = \begin{cases} 
      0 & \mathrm{if}\  \mathrm{pn}(i)\ \mathrm{is}\ \infty \\
       1 - \frac{\mathrm{pn}(i)}{1+\sum_{j \in \mathcal{I}  \wedge \mathrm{pn}(j)\neq\infty}\mathrm{pn}(j)} & \mathrm{otherwise} \\
   \end{cases}\\   
\end{equation}

The unique property of \pnsum\ is that its behavior is far more strongly correlated with the branching factor of the parent node. than in the other tested formulas. This, depending on the game and the test setting, may turn out as a benefit or a liability.

\subsubsection{\cpn\ constant tuning}

Previous research regarding PN-MCTS was somewhat based on the assumption that the \cpn constant is universal, and similarly to the MCTS $C$ constant, the same value can be used across a variety of games.
Thus, the relation between \cpn\ and each tested game was not properly studied.
As our approach introduces alternative functions to serve as UCT-PN formulas, the need for game- and formula-based tuning seems even more important.

Although the experiments and associated discussion are presented in Sec.~\ref{sec:2presults}, we think it is worth stating here, as an essential part of the description of the algorithm that indeed, the \cpn value has a crucial impact on the performance of GPN-MCTS.
Moreover, the peak performance may occur at every point of the spectrum of tested values, as well as the peak for different UCT-PN formulas may be achieved by different \cpn\ values for the same game.






\subsubsection{Backpropagation optimization}\label{sec:backpropoptim}

All three UCT-PN bias formulas share the same characteristic: their value for a given node depends on all children of the same parent.
In case of \pnmax\ and \pnsum, these are factors that can be associated with the parent node and simply stored there as precomputed values, but for \pnrank\ the formula is more complicated and strongly benefits computing UCT-PN value for all children at once.

This, however, complicates the usual backpropagation step of PNS. Originally, it is a cheap operation requiring an update of a parent value if any child value was modified.
Now, apart from just a proof number, we also need to take into account a UCT-PN value, and thus, all children of the parent should be updated, which is significantly more expensive, especially for \pnrank.

However, we can at least partially mitigate that cost. The main observation is that not all values that would be changed during backpropagation will actually be called by the UCT-PN during the MCTS selection step.
Thus, in our implementation, we optimized this step by setting the \emph{needRecalc} flag on a node with updated proof number, and do not precompute UCT-PN value for this parent nor its siblings. UCT-PN formula is computed in a call-by-need manner, only if a \emph{needRecalc} node is encountered during the selection phase (and after, the flag is set to false).

\subsection{Mobility-based Initialization}
Most classic improvements of PNS are focused on handling the issue of memory consumption when attempting to solve the game tree \cite{breuker01b,nagai98}, so they have no application when proof numbers are used as a selection bias inside the game-playing algorithm.

However, the \textit{mobility initialization} enhancement \cite{vandenHerik2008} can be straightforwardly applied to GPN-MCTS.
The idea is to initialize unknown leaf nodes in a more elaborate way than the one described in Section~
\ref{sec:pns}.
In an AND node, the proof number can be set to the number of legal moves in this node.
(In PNS, the disproof numbers are initialized that way in OR nodes.)
This optimization improves the quality of (dis)proof numbers, as it works as one-step lookahead for computing the estimated sizes of the subtrees to prove.


\subsection{Score Bounded GPN-MCTS}

In practical implementations, algorithms like MCTS are nearly never applied in a vanilla format. 
Usually, the resulting algorithm consists of the union of a few general enhancements plus game-dependent improvements.
For this reason, one of our goals was to perform tests of the proposed GPN-MCTS using a well-established advanced algorithm setup and see if the improved results carry over. We decided to put as our baseline an MCTS version with two improvements. One is simply a standard technique of tree reuse \cite{mctssurvey}, and the other is Score Bounded MCTS \cite{cazenave2011ScoreBoundedMCTS}.

Score-Bounded Monte-Carlo Tree Search (SB MCTS) is an extension of MCTS Solver \cite{winands08b} that can handle draws, and is generalized to games with many outcomes, as well as games with more than two players.
Subtree-solving is probably the only line of MCTS enhancements that may be considered obligatory, as it comes with nearly no drawbacks.
It is easy to implement, has a negligible computational cost, and more often than not increases the playing strength of an agent when applied to zero-sum games.

Proof-Number Search and Score-Bounded MCTS are based on a similar premise; in both cases, we are tracking which subtrees of the search tree are solved.
On the one hand, PNS stores data allowing it to predict which subtrees can be solved with less effort, the information SB MCTS is lacking.
On the other hand, SB MCTS requires only two values (lower bound, upper bound) per tree node to store guaranteed payoffs, regardless of the number of possible outcomes in the game.
For the purpose of biasing UCB in PN-MCTS, this would require a number of proof trees linear with respect to possible game outcomes.
Such an approach was proposed in \cite{Kowalski2024ProofNumber}, introducing a second PN tree to handle draws.
(Note that keeping just \emph{attracting outcome} as in Multiple-Outcome PNS \cite{saffidine2012multiple} does not work when applied to dynamic in-game search.)

We argue that using a well-established and more general solution, such as the Score Bounded extension, is a better option than further extending PN-MCTS into overlapping tasks of skipping solved subtrees and biasing a final move selection, as intended by \cite{Kowalski2024ProofNumber}. Although PN trees can be used that way, it is inefficient, and the resulting code is less manageable. 

The results (shown in Sec.~\ref{sec:sbresults}) confirm that GPN-MCTS works well when combined with the Score Bound method, 
and there is no significant loss of quality between results on GPN-MCTS vs.\ MCTS and Score Bounded GPN-MCTS vs.\ Score Bounded MCTS.

\subsection{The GPN-MCTS Algorithm}

The pseudocode for a single iteration of GPN-MCTS is shown in Alg.~\ref{alg:gpnmcts}.
The overall frame of the algorithm follows the MCTS definition from \cite{mctssurvey}, and uses the same terminology when possible. For example, $\Call{DefaultPolicy}{s}$ encodes the standard simulation for the given game state to a terminal state.

A flag $\mathit{needRecalc}$ of a node is the one introduced by optimization described in Section~\ref{sec:backpropoptim}, and is set to false inside the $\Call{UpdateChildrenPNScores}{ }$ procedure.
The details of this procedure are omitted, as it just calculates selection bias values of the children nodes for the node's moving player, according to the formulas from Section~\ref{sec:pnformulas}.

$\Call{BestUCTPNChild}{}$ returns a child that maximizes the value of the UCT-PN Formula (\ref{eq:uctpn}), with a comment that other bias functions, instead of \pnrank\ might be encoded there.

The $\Call{Backup}{v_l, \Delta}$ function consists of two separate parts. One is the standard MCTS backpropagation. The other is based on \emph{updateAncestors} procedure (c.f.\ \cite{KishimotoW0S12}) to backup proof numbers in an optimized way (early stop when no change is detected).

Finally, $\Call{UpdateProofNumber}{p}$, updates the proof number in a node for a given player according to the PNS \emph{setProofAndDisproofNumbers} procedure.
The main differences here is that we update only proof numbers and that the OR/AND node distinction is based on whether $p$ is a player performing a move at that node or not. 
A $\mathit{proofWinner}$ in a leaf node is either $\mathrm{Unknown}$ if the associated game state is not terminal, or the player that won the game otherwise.

\begin{algorithm}[htb!]\footnotesize
\begin{algorithmic}[1]
\Require{$v_0$ -- current root node of the GPN-MCTS tree}
\Function{GPN-MCTS-Iteration}{$v_0$ }
\State $v_l \gets \Call{TreePolicy}{v_0}$ 
\State $\Delta \gets \Call{DefaultPolicy}{v_l.\Call{State}{ }}$
\State $\Call{Backup}{v_l, \Delta}$
\EndFunction
\\
\Require{$v$ -- root node of the GPN-MCTS tree}
\Function{TreePolicy}{$v$}
\While{$v.\Call{State}{ }.\Call{IsNotTerminal}{ }$}
\LineIf{$v.\mathit{needRecalc}$}{$v.\Call{UpdateChildrenPNScores}{ }$ }
\If{$v.\Call{IsNotFullyExpanded}{ }$}
 \State \Return $\Call{Expand}{v}$
\Else
\State $v \gets v.\Call{BestUCTPNChild}{ }$
\EndIf
\EndWhile
\State \Return $v$
\EndFunction
\\
\Require{$v$ -- last node of iteration in GPN-MCTS tree}
\Function{Expand}{$v$}
\State $a \gets \Call{RandomElement}{v.\Call{UntriedMoves}{ }}$
\State $v' \gets \Call{CreateNode}{v.\Call{State}{ }.\Call{Apply}{a}}$
\LineForAll{$p \in\Call{Players}{ }$}{$v'.\Call{UpdateProofNumber}{p}$}
\State $v.\Call{AddNode}{v', a}$
\State \Return $v'$ 
\EndFunction
\\
\Require{$\mathit{v_l}$ -- last node of iteration in GPN-MCTS tree}
\Require{$\Delta$ --  final scores for each player of iteration}
\Function{Backup}{$v_l, \Delta$}
\State $v \gets \mathit{v_l}$
\While{$v \neq \mathbf{None}$}
   \State $v.\mathit{scoreSum} \gets v.\mathit{scoreSum} + \Delta[leaf.\mathit{player}]$
   \State $v.\mathit{iterations} \gets v.\mathit{iterations} + 1$
   \State $v \gets v.\Call{Parent}{ }$
\EndWhile
\ForAll{$p \in\Call{Players}{ }$}
   \State $v \gets \mathit{v_l}$ 
   \State $c \gets \mathbf{True}$
   \While{$c$ $\mathbf{and}$ $v \neq \mathbf{None}$}
   \State $c \gets v.\Call{UpdateProofNumber}{p}$ 
   \LineIf{$c$}{$v.\mathit{needRecalc} \gets \mathbf{true}$}
   \State $v \gets v.\Call{Parent}{ }$
   \EndWhile
\EndFor
\EndFunction
\\
\Require{$\mathit{p}$ -- player for whom the update takes place}
\Function{Node.UpdateProofNumber}{$p$}
\If{$\Call{IsNotExpanded}{ }$} 
\LineIfElseElse{$\mathit{proofWinner}=\mathbf{Unknown}$}{$\mathit{proofNumber} \gets 1$}
{$\mathit{proofWinner}=p$}{$\mathit{proofNumber} \gets 0$} 
{$\mathit{proofNumber} \gets \infty$} 
\State \Return $\mathbf{True}$
\EndIf
\State $\mathit{oldProof} \gets \mathit{proofNumber}$
\If{$p=\mathit{nodePlayer}$} \Comment{PNS OR node}
\State $\mathit{proofNumber} \gets \infty$
\DLineForAll{$\mathit{child}\in \Call{children}{ }$}{$\mathit{proofNumber} \gets \Call{min}{\mathit{proofNumber},  child.\mathit{proofNumber}}$}
\Else \Comment{PNS AND node}
\State $\mathit{proofNumber} \gets 0$
\DLineForAll{$\mathit{child}\in \Call{children}{ }$}{$\mathit{proofNumber} \gets \mathit{proofNumber} + child.\mathit{proofNumber}$}
\EndIf
\State \Return $\mathit{proofNumber}\neq\mathit{oldProof}$
\EndFunction

\end{algorithmic}
\caption{The GPN-MCTS Algorithm}\label{alg:gpnmcts}
\end{algorithm}

\section{Experiments}\label{sec:Exp}


The experiments have been conducted using the Ludii general game-playing system, which provides an environment for developers to test their implementation of general game-playing agents \cite{Piette_2020_Ludii}.
It was chosen as it contains over 1,000 games described in its game description language, and implementations of various standard algorithms and enhancements (such as several variants of MCTS), with a single, unified API for the development of AI agents.

The presented GPN-MCTS algorithm has been implemented as an enhancement of the agents available in Ludii.\footnote{The source code is available in the appendix and will be made public after the review process.} It is consistent with the latest publicly available version of the framework  (ver.\ 1.3.14) and will be proposed as a pull request after the reviewing process to the official repository of the project.\footnote{\url{https://github.com/Ludeme/Ludii}}
Two versions of GPN-MCTS are available: with Score Bounded enhancement and without. 
If not stated otherwise, the experiments are performed by playing with Score Bounded GPN-MCTS with tree reuse against Score Bounded MCTS with tree reuse.
For both agents, the MCTS $C$ parameter is set to $\sqrt{2}$.
Player positions are swapped in all tests so that the agents play both sides equally often, and draws count as half wins.  
The experiments were performed on different machines; however, for any game, all results regarding this game were computed using the same hardware, which makes them comparable.
For every result, if any error margins are presented, they represent a 95\% confidence interval.

\begin{table*}[!ht]\small\renewcommand{\arraystretch}{1.1}
\caption{The results of GPN-MCTS versus MCTS. Variant including SB means both algorithms use Score Bounded and tree reuse, otherwise it is just tree reuse. (500 games, 1s per turn).}
\label{tab:2presults}
\begin{center}\begin{tabular}{l||c|c|c|c|c|c}
\toprule
\multirow{2}{*}{Variant}& \multicolumn{6}{c}{$C_{pn}$ value} \\ 
 & 0.0 & 0.1 & 0.5 & 1.0 & 2.0 & 5.0 \\ \hline 

 \multicolumn{7}{c}{Ataxx} \\ 
 \pnrank & $45.9\pm4.37$  & $58.0\pm4.32$ & $82.3\pm3.33$  & $87.0\pm2.95$ & $89.4\pm2.70$  &  $\textbf{89.5}\pm2.67$  \\ 
\pnrank+SB & $50.1\pm4.38$ &  $60.6\pm4.27$   &  $85.3\pm3.09$ & $89.2\pm2.72$  & $91.4\pm2.46$    & $\textbf{92.0}\pm2.36$ \\ 
\pnmax+SB & $49.2\pm4.38$  & $65.3\pm4.17$  &  $\textbf{92.4}\pm2.33$  & $92.3\pm2.33$   & $91.0\pm2.51$   &      $89.1\pm2.71$  \\ 
\pnsum+SB & $45.6\pm4.36$  &  $56.5\pm4.33$  &  $55.7\pm4.34$ &  $56.2\pm4.34$  &   $61.6\pm4.25$ &   $\textbf{71.2}\pm3.93$ \\ 
\hline

\multicolumn{7}{c}{Awari } \\ 
\pnrank+SB & $48.1\pm3.97$ & $71.0\pm3.53$ & $\textbf{74.3}\pm3.32$ & $62.2\pm3.63$ & $50.4\pm3.92$ & $43.2\pm4.21$ \\
\pnmax+SB  & $50.9\pm3.95$ & $\textbf{78.2}\pm3.21$ & $69.8\pm3.50$ & $56.3\pm4.28$ & $45.5\pm4.27$ & $45.1\pm4.25$     \\ 
\pnsum+SB  & $49.8\pm4.04$ & $62.2\pm3.73$ & $78.8\pm3.28$ & $\textbf{79.7}\pm3.07$ & $77.2\pm3.34$ & $67.9\pm3.77$     \\
\hline

\multicolumn{7}{c}{Knightthrough } \\
\pnrank & $48.0\pm4.38$ & $50.6\pm4.39$ & $52.2\pm4.38$ & $52.8\pm4.38$  & $62.6\pm4.25$ & $\textbf{74.0}\pm3.85$ \\ 
\pnrank+SB & $53.2\pm4.38$ & $45.8\pm4.37$ & $46.2\pm4.37$ & $56.3\pm4.33$ & $54.8\pm4.37$ & $\textbf{59.2}\pm4.31$   \\ 
\pnmax+SB & $46.8\pm4.38$ & $56.6\pm4.35$ & $50.8\pm4.39$ & $62.8\pm4.24$ & $\textbf{63.2}\pm4.21$ & $54.4\pm4.37$     \\ 
\pnsum+SB & $49.0\pm4.39$ & $50.0\pm4.39$ & $56.0\pm4.36$ & $52.0\pm4.38$ & $55.8\pm4.36$ & $\textbf{60.0}\pm4.30$     \\ 
\hline

\multicolumn{7}{c}{Lines of Action ($7\times 7$)} \\ 
\pnrank & $52.1\pm4.38$ & $67.4\pm4.11$ & $85.0\pm3.13$ & $87.0\pm2.95$ & $\textbf{88.4}\pm2.81$ & $80.2\pm3.49$ \\ 
\pnrank+SB & $47.6\pm4.37$ & $67.1\pm4.12$ & $82.8\pm3.31$ & $91.3\pm2.47$ & $\textbf{91.8}\pm2.39$ & $83.6\pm3.25$ \\ 
\pnmax+SB   & $52.6\pm4.38$ & $66.3\pm4.14$ & $81.3\pm3.42$ & $\textbf{84.3}\pm3.14$ &  $68.6\pm4.06$ & $65.8\pm4.13$     \\ 
\pnsum+SB   & $54.6\pm4.37$ & $55.1\pm4.36$ & $60.9\pm4.28$ & $64.9\pm4.18$  & $67.9\pm4.08$ & $\textbf{69.3}\pm4.04$     \\
\hline
\multicolumn{7}{c}{Lines of Action ($8\times 8$) } \\ 
\pnrank+SB & $51.8\pm4.38$ & $63.6\pm4.22$ & $83.8\pm3.23$ & $\textbf{87.4}\pm2.91$ & $81.9\pm3.37$ & $55.2\pm4.35$ \\ 
\pnmax+SB  & $50.8\pm4.39$ & $55.8\pm4.36$ & $68.8\pm4.07$ & $\textbf{82.5}\pm3.33$ & $63.7\pm4.21$ & $25.9\pm3.83$     \\ 
\pnsum+SB  & $48.0\pm4.38$ & $54.5\pm4.33$ & $52.4\pm4.38$ & $55.6\pm4.41$ & $55.5\pm4.75$ & $\textbf{64.3}\pm4.48$     \\
\hline

\multicolumn{7}{c}{Los Alamos chess} \\ 
\pnrank & $53.2\pm4.10$ & $71.9\pm3.76$  & $\textbf{85.5}\pm2.94$  &  $82.4\pm3.22$ & $80.5\pm3.36$  &   $80.5\pm3.39$    \\
\pnrank+SB & $47.3\pm4.16$     &  $71.7\pm3.73$ & $81.0\pm3.33$ & $\textbf{84.5}\pm3.02$  &  $80.5\pm3.38$   &   $79.3\pm3.43$      \\ 
\pnmax+SB &  $48.7\pm4.16$ &  $71.1\pm3.74$ & $73.4\pm3.73$   & $76.5\pm3.52$ &  $\textbf{80.5}\pm3.35$  & $78.7\pm3.45$ \\ 
\pnsum+SB & $49.7\pm4.11$   & $60.3\pm4.10$   & $64.5\pm3.95$ & $72.7\pm3.75$   &  $\textbf{77.5}\pm3.46$ & $77.0\pm3.54$ \\ \hline

 \multicolumn{7}{c}{Minishogi} \\ 
 \pnrank & $47.4\pm4.38$ &  $55.0\pm4.37$   & $67.8\pm4.17$  & $\textbf{69.2}\pm4.05$ &    $56.8\pm4.35$ 
    &  $45.6\pm4.37$    \\
\pnrank+SB & $45.2\pm4.37$ &   $54.4\pm4.37$ & $\textbf{67.8}\pm4.10$  & $66.2\pm4.15$   &   $63.8\pm4.22$      &  $46.0\pm4.37$   \\
\pnmax+SB & $52.4\pm4.38$ &  $49.6\pm4.39$    & $\textbf{51.2}\pm4.39$ & $34.6\pm4.17$ & $31.6\pm4.08$ & $36.2\pm4.22$ \\ 
\pnsum+SB &  $49.6\pm4.39$   & $52.4\pm4.38$ & $57.0\pm4.34$ & $\textbf{59.4}\pm4.31$ & $51.8\pm4.38$ & $48.8\pm4.39$ \\ \hline

\multicolumn{7}{c}{Reach Chess} \\ 
\pnrank &$44.4\pm4.36$  & $58.6\pm4.32$  &  $64.8\pm4.19$  & $66.2\pm4.15$ & $71.8\pm3.95$  &  $\textbf{76.8}\pm3.70$ \\ 
\pnrank+SB & $50.2\pm4.39$ & $62.0\pm4.26$ & $64.6\pm4.20$ & $72.2\pm3.93$ & $\textbf{79.0}\pm3.57$ & $78.0\pm3.64$ \\ 
\pnmax+SB & $49.2\pm4.39$ & $58.8\pm4.32$ & $\textbf{70.2}\pm4.01$ & $65.2\pm4.18$  & $69.8\pm4.03$ &  $55.2\pm4.36$  \\
\pnsum+SB & $46.2\pm4.37$     &  $57.4\pm4.34$     & $59.8\pm4.30$   & $63.6\pm4.22$ & $64.4\pm4.20$ &  $\textbf{68.2}\pm4.09$    \\
\hline

\multicolumn{7}{c}{Reversi} \\ 
\pnrank & $48.7\pm4.28$ & $65.1\pm4.11$ & $\textbf{71.6}\pm3.89$  & $49.9\pm4.32$   & $43.4\pm4.33$  &  $42.5\pm4.32$  \\ 
\pnrank+SB & $50.0\pm4.32$ & $\textbf{71.0}\pm3.87$   & $69.4\pm3.99$  & $53.2\pm4.34$ & $42.4\pm4.27$  & $37.2\pm4.22$ \\
\pnmax+SB & $50.0\pm4.24$ & $\textbf{63.1}\pm4.16$ & $59.0\pm4.26$ & $45.2\pm4.30$ & $36.2\pm4.18$ & $35.0\pm4.15$    \\ 
\pnsum+SB & $49.5\pm4.28$ & $55.0\pm4.27$ & $60.7\pm4.21$ & $71.7\pm3.86$ & $\textbf{76.7}\pm3.64$ & $64.1\pm4.14$    \\
\hline

\multicolumn{7}{c}{Skirmish} \\ 
\pnrank+SB & $50.6\pm4.16$ & $\textbf{62.4}\pm4.07$ & $62.3\pm4.11$ & $61.7\pm4.17$ & $60.8\pm4.17$ & $55.8\pm4.29$ \\
\pnmax+SB  & $51.5\pm4.20$ & $65.1\pm4.00$ & $\textbf{65.3}\pm4.05$ & $58.0\pm4.17$ & $56.2\pm4.26$ & $39.0\pm4.18$    \\ 
\pnsum+SB  & $50.5\pm4.23$ & $49.9\pm4.19$ & $56.4\pm4.11$ & $55.3\pm4.12$ & $63.4\pm4.38$ & $\textbf{63.8}\pm4.03$    \\
\hline

\multicolumn{7}{c}{Surakarta} \\ 
\pnrank+SB & $49.5\pm4.34$  & $60.5\pm4.23$ &  $\textbf{60.6}\pm4.25$ &  $59.1\pm4.26$  & $46.7\pm4.33$     &   $15.2\pm3.11$        \\
\pnmax+SB & $48.6\pm4.36$ &$65.1\pm4.13$  & $\textbf{82.0}\pm3.33$  &  $68.4\pm4.05$  & $44.4\pm4.32$   &   $22.4\pm3.67$      \\ 
\pnsum+SB &  $47.8\pm4.35$  & $52.5\pm4.63$  & $54.3\pm4.50$ &  $50.9\pm4.35$   &  $54.6\pm4.34$ &   $\textbf{57.7}\pm4.27$   \\ 
\bottomrule
\end{tabular}
\\[2cm] 
\end{center}

\end{table*}

\subsection{UCT-PN Bias Formulas and \cpn}\label{sec:2presults}

The main experiments were conducted on a set of two-player, zero-sum board games, including the games used for experiments in \cite{Kowalski2024ProofNumber}.
Our aim is to test the behavior of each bias formula variant (\pnrank, \pnmax, \pnsum) and the influence of \cpn\ constant on the improvements over the baseline agent.
We tested $\cpn\in \{0.0, 0.1, 0.5, 1.0, 2.0, 5.0\}$ to be consistent with the previous research regarding PN-MCTS. Although this range of parameters may be insufficient to provide the exact highest winrate available for a given game and UCT-PN variant, it provides a good estimation of them, as well as of the general behavior of the winrate function for these settings.
Each of the tests consists of 500 games with 0.5 second per turn. The results are presented in Table \ref{tab:2presults}.

GPN-MCTS achieved 80\% win rate against the Score Bounded MCTS on 8 out of the 11 tested board games, in two cases (Ataxx and Lines of Action) even reaching 90\%.
For the remaining tested games, the results are also confident wins, with the lowest best score 63.2\% obtained for Knighttrough.

The first observation is that the best \cpn\ values greatly differ for various games.
For some games, the best value was the lowest one tested, and for some, the largest one (which suggests that even the better \cpn\ values can potentially be found outside of the tested range).
Also, when looking at most of the games, the spread of the winrate between the best and worst choice of \cpn\ is vast (in extreme case, for Lines of Action $8\times8$, the winrate degraded from 82.5\% for $\cpn=1$ to 25.9\% for $\cpn=5$).
This is a strong indicator that picking a value that behaves best on average is not a good idea, and the results obtained that way may be very far from the optimal ones.

The second observation is based on comparing the behavior and results for the three tested bias formulas.
There is no clear winner on which of the formulas is the best. Especially as often the peak winrates are not far from each other, and local \cpn\ tuning can possibly reverse the results. (And sometimes, especially for \pnsum\ which has a tendency to have peaks for larger \cpn\ values, it may be even further outside of the tested parameter range.)
Generally, \pnrank\ seems like a safe choice, obtaining the best results, or relatively close to best, for most games. In some cases, like for Minishogi, this bias formula shows a clear advantage over the others.
In others, e.g.,\ for Surakarta, \pnmax\ is clearly better with over 20 percent point advantage.
Although for many games \pnsum\ has lower results, there are also cases (Reversi) when it performs the best out of three.

Summarizing, if the game is susceptible to PN-based exploration, then using any of the formulas should lead to improvement, but the actual amount of this improvement depends on the particular pick and may differ significantly.

\subsection{Influence of Score Bounded on GPN-MCTS}\label{sec:sbresults}

For some of the games and the \pnrank\ formula, we run experiments without the Score Bounded enhancement (for both GPN-MCTS and MCTS).
Thus, we can analyze if the PN-based works as well as the vanilla algorithm as with a union with other improvements.
The potential problem of any improvement to any complex AI algorithm is that although it works standalone, its benefits are degraded when applied with other improvements. 
Also, in more real-world scenarios, we cannot expect our opponent to be a basic implementation, so the enhancements should be able to show improvements also against a more advanced opponent.

The results of our comparison show that, in general, improving both GPN-MCTS and the opponent with the Score Bounded extension keeps the win rates in similar ranges.
Note, however, that the PN-MCTS, as its core goal is to bias exploration towards potentially fast-to-prove nodes, is especially suited to work well with the Score Bounded extension. 
This may explain why often the results including SB have a tendency to be (slightly) better than for vanilla MCTS.

\subsection{Overhead}

The requirement of maintaining PNS-related structures on top of the standard MCTS tree implies that PN-MCTS is usually slower than the pure MCTS. 
Thus, it will generally perform fewer iterations within any given time budget.
For this reason, all experiments in this paper use time-based budgets, which we consider a fair comparison, as simulation-based budgets ignore the factor of computation overhead.
GPN-MCTS is developed as an extension of the MCTS implementation provided by the Ludii system to ensure that any difference in performance is solely due to the implementation of the proposed enhancement.

To measure the overhead, we included $\cpn=0$ in the results of Table \ref{tab:2presults}. This winrate value represents the match between two Score Bounded MCTS algorithms (in terms of behavior), but with one spending additional computation time on managing proof numbers and UCT-PN values.
Most of these win rates are within the confidence interval distance to 50\%, which means that the computational overhead of GPN-MCTS seems to be small enough not to affect the expected results. 

\section{Conclusion}\label{sec:conclusion}


The experimental results show that GPN-MCTS is a rather impactful MCTS enhancement, and we argue that it has all the reasons to be considered as a staple improvement to implement alongside Solver/MAST/RAVE and other classic developments.

It is relatively simple to implement, based on a classic, well-described algorithm.
It pairs with obligatory Score Bounded MCTS enhancement.
It is easy to test whether its application will be beneficial for a given game -- comparing an algorithm without extension with any bias formula and one or two small \cpn\ values should give the right indication.
And the potential gains can range from a trustworthy 60\% win up to a 90\% decisive victory.


In this paper, we focused our experiments on two-player games, despite the fact that GPN-MCTS can actually handle games with more players. Although multi-player games are an interesting challenge, so far integrating solving capabilities in  MCTS has led only to limited improvements \cite{Nijssen_2011_Enhancements}. Therefore, we have considered the application of GPN-MCTS in these domains as future research.

Our preliminary experiments have indicated that there are games where GPN extension makes no visible impact (Pentago, Pentalah) or even worsens the results (Connect Four, Diagonals). This ought to be expected, as Proof-Number Search does not always work when applied in a game-agnostic manner. PNS takes advantage of situations when there are deep, narrow winning paths. For some games, introducing domain knowledge \cite{KishimotoW0S12} is required to shape the search tree in such a way that narrow and forced paths emerge. Testing if transferring such knowledge to GPN-MCTS framework is possible and it will positively influence results for such games is one of the promising paths for future work.





\begin{ack}

This article is based on the work of COST Action CA22145 -- GameTable, supported by COST (European Cooperation in Science and Technology).

This research was supported in part by the National Science Centre, Poland, under project number 2021/41/B/ST6/03691 (Jakub Kowalski). 




\end{ack}



\bibliography{bibliography,Dennis-Soemers-Bib}

\begin{thebibliography}{28}
\providecommand{\natexlab}[1]{#1}
\providecommand{\url}[1]{\texttt{#1}}
\expandafter\ifx\csname urlstyle\endcsname\relax
  \providecommand{\doi}[1]{doi: #1}\else
  \providecommand{\doi}{doi: \begingroup \urlstyle{rm}\Url}\fi

\bibitem[Allis et~al.(1994)Allis, van~der Meulen, and van~den Herik]{allis94}
L.~V. Allis, M.~van~der Meulen, and H.~J. van~den Herik.
\newblock {Proof-Number Search}.
\newblock \emph{Artificial Intelligence}, 66\penalty0 (1):\penalty0 91--123, 1994.

\bibitem[Arneson et~al.(2010)Arneson, Hayward, and Henderson]{arneson10}
B.~Arneson, R.~B. Hayward, and P.~Henderson.
\newblock {Monte Carlo Tree Search in Hex}.
\newblock \emph{IEEE Transactions on Computational Intelligence and AI in Games}, 2\penalty0 (4):\penalty0 251--258, 2010.

\bibitem[Auer et~al.(2002)Auer, Cesa-Bianchi, and Fischer]{auerfinit02}
P.~Auer, N.~Cesa-Bianchi, and P.~Fischer.
\newblock {Finite-Time Analysis of the Multiarmed Bandit Problem}.
\newblock \emph{Machine Learning}, 47\penalty0 (2--3):\penalty0 235--256, 2002.

\bibitem[Baier and Winands(2015)]{baier2015}
H.~Baier and M.~H.~M. Winands.
\newblock {MCTS-Minimax Hybrids}.
\newblock \emph{IEEE Transactions on Computational Intelligence and AI in Games}, 7\penalty0 (2):\penalty0 167--179, 2015.
\newblock \doi{10.1109/TCIAIG.2014.2366555}.

\bibitem[Bj\"{o}rnsson and Finnsson(2009)]{bjornsson09}
Y.~Bj\"{o}rnsson and H.~Finnsson.
\newblock {CadiaPlayer}: {A} {Simulation-Based} {General Game Player}.
\newblock \emph{IEEE Transactions on Computational Intelligence and AI in Games}, 1\penalty0 (1):\penalty0 4--15, 2009.

\bibitem[Breuker et~al.(2001)Breuker, Uiterwijk, and van~den Herik]{breuker01b}
D.~M. Breuker, J.~W.~H.~M. Uiterwijk, and H.~J. van~den Herik.
\newblock The {PN$^2$}-search algorithm.
\newblock In H.~J. van~den Herik and B.~Monien, editors, \emph{Advances in Computer Games 9}, pages 115--132. Universiteit Maastricht, Maastricht, The Netherlands, 2001.

\bibitem[Browne et~al.(2012)Browne, Powley, Whitehouse, Lucas, Cowling, Rohlfshagen, Tavener, Perez, Samothrakis, and Colton]{mctssurvey}
C.~B. Browne, E.~Powley, D.~Whitehouse, S.~M. Lucas, P.~I. Cowling, P.~Rohlfshagen, S.~Tavener, D.~Perez, S.~Samothrakis, and S.~Colton.
\newblock {A Survey of {M}onte {C}arlo {Tree Search} Methods}.
\newblock \emph{{IEEE} Transactions on Computational Intelligence and {AI} in Games}, 4\penalty0 (1):\penalty0 1--43, 2012.

\bibitem[Cazenave and Saffidine(2011)]{cazenave2011ScoreBoundedMCTS}
T.~Cazenave and A.~Saffidine.
\newblock {Score Bounded Monte-Carlo Tree Search}.
\newblock In \emph{Computers and Games}, volume 6515 of \emph{LNCS}, pages 93--–104, 2011.

\bibitem[Chaslot et~al.(2008)Chaslot, Winands, van~den Herik, Uiterwijk, and Bouzy]{chaslot08}
G.~M.~J.-B. Chaslot, M.~H.~M. Winands, H.~J. van~den Herik, J.~W.~H.~M. Uiterwijk, and B.~Bouzy.
\newblock {Progressive Strategies for {Monte-Carlo Tree Search}}.
\newblock \emph{New Mathematics and Natural Computation}, 4\penalty0 (3):\penalty0 343--357, 2008.

\bibitem[Coulom(2007)]{coulom06}
R.~Coulom.
\newblock {Efficient Selectivity and Backup Operators in {M}onte-{C}arlo {Tree Search}}.
\newblock In \emph{Computers and Games (CG 2006)}, volume 4630 of \emph{Lecture Notes in Computer Science}, pages 72--83, 2007.

\bibitem[Doe et~al.(2022)Doe, Winands, Soemers, and Browne]{Doe_2022_Combining}
E.~Doe, M.~H.~M. Winands, D.~J. N.~J. Soemers, and C.~Browne.
\newblock Combining {M}onte-{C}arlo tree search with proof-number search.
\newblock In \emph{Proceedings of the 2022 IEEE Conference on Games}, pages 206--212, 2022.

\bibitem[Kishimoto et~al.(2012)Kishimoto, Winands, M{\"{u}}ller, and Saito]{KishimotoW0S12}
A.~Kishimoto, M.~H.~M. Winands, M.~M{\"{u}}ller, and J.-T. Saito.
\newblock {Game-Tree Search Using Proof Numbers: The First Twenty Years}.
\newblock \emph{ICGA Journal}, 35\penalty0 (3):\penalty0 131--156, 2012.

\bibitem[Kocsis and Szepesv{\'a}ri(2006)]{Kocsis_2006_Bandit}
L.~Kocsis and C.~Szepesv{\'a}ri.
\newblock Bandit based {M}onte-{C}arlo planning.
\newblock In J.~F{\"u}rnkranz, T.~Scheffer, and M.~Spiliopoulou, editors, \emph{Machine Learning: ECML 2006}, volume 4212 of \emph{Lecture Notes in Computer Science}, pages 282--293. Springer, Berlin, Heidelberg, 2006.

\bibitem[Kowalski et~al.(2024)Kowalski, Doe, Winands, G\'{o}rski, and Soemers]{Kowalski2024ProofNumber}
J.~Kowalski, E.~Doe, M.~H.~M. Winands, D.~G\'{o}rski, and D.~J. N.~J. Soemers.
\newblock {Proof Number Based Monte-Carlo Tree Search}.
\newblock \emph{IEEE Transactions on Games}, 17\penalty0 (1):\penalty0 148--157, 2024.

\bibitem[Lanctot et~al.(2014)Lanctot, Winands, Pepels, and Sturtevant]{LanctotWPS14}
M.~Lanctot, M.~H.~M. Winands, T.~Pepels, and N.~R. Sturtevant.
\newblock {Monte Carlo Tree Search with Heuristic Evaluations using Implicit Minimax Backups}.
\newblock In \emph{2014 {IEEE} Conference on Computational Intelligence and Games, {CIG} 2014}, pages 341--348, 2014.

\bibitem[Lorentz(2008)]{Lorentz08}
R.~J. Lorentz.
\newblock {Amazons Discover Monte-Carlo}.
\newblock In \emph{Computers and Games (CG 2008)}, volume 5131 of \emph{Lecture Notes in Computer Science}, pages 13--24, 2008.

\bibitem[Nagai(1998)]{nagai98}
A.~Nagai.
\newblock A new {AND/OR} tree search algorithm using proof number and disproof number.
\newblock In \emph{Proceedings of Complex Games Lab Workshop}, pages 40--45. ETL, Tsukuba, Japan, 1998.

\bibitem[Nijssen and Winands(2011)]{Nijssen_2011_Enhancements}
J.~A.~M. Nijssen and M.~H.~M. Winands.
\newblock Enhancements for multi-player {M}onte-{C}arlo tree search.
\newblock In H.~J. van~den Herik, H.~Iida, and A.~Plaat, editors, \emph{Computers and Games (CG 2010)}, volume 6515 of \emph{Lecture Notes in Computer Science}, pages 238--249. Springer Berlin Heidelberg, 2011.

\bibitem[Piette et~al.(2020)Piette, Soemers, Stephenson, Sironi, Winands, and Browne]{Piette_2020_Ludii}
{\'E}.~Piette, D.~J. N.~J. Soemers, M.~Stephenson, C.~F. Sironi, M.~H.~M. Winands, and C.~Browne.
\newblock Ludii -- the ludemic general game system.
\newblock In G.~D. Giacomo, A.~Catala, B.~Dilkina, M.~Milano, S.~Barro, A.~Bugarín, and J.~Lang, editors, \emph{Proceedings of the 24th European Conference on Artificial Intelligence (ECAI 2020)}, volume 325 of \emph{Frontiers in Artificial Intelligence and Applications}, pages 411--418. IOS Press, 2020.

\bibitem[Randall et~al.(2024)Randall, M\"{u}ller, Wei, and Hayward]{Randall2024}
O.~Randall, M.~M\"{u}ller, T.-H. Wei, and R.~Hayward.
\newblock Expected work search: {C}ombining win rate and proof size estimation.
\newblock In \emph{Proceedings of the Thirty-Third International Joint Conference on Artificial Intelligence}, IJCAI '24, pages 7003 -- 7011, 2024.

\bibitem[Saffidine and Cazenave(2012)]{saffidine2012multiple}
A.~Saffidine and T.~Cazenave.
\newblock {Multiple-outcome Proof Number Search}.
\newblock In \emph{ECAI}, pages 708--713. 2012.

\bibitem[Saito and Winands(2010)]{Saito10}
J.-T. Saito and M.~H.~M. Winands.
\newblock {Paranoid Proof-Number Search}.
\newblock In \emph{Proceedings of the Computational Intelligence and Games Conference (CIG'10)}, pages 203--210, 2010.

\bibitem[Silver et~al.(2017)Silver, Schrittwieser, Simonyan, Antonoglou, Huang, Guez, Hubert, Baker, Lai, Bolton, Chen, Lillicrap, Hui, Sifre, van~den Driessche, Graepel, and Hassabis]{Silver2017mastering}
D.~Silver, J.~Schrittwieser, K.~Simonyan, I.~Antonoglou, A.~Huang, A.~Guez, T.~Hubert, L.~Baker, M.~Lai, A.~Bolton, Y.~Chen, T.~Lillicrap, F.~Hui, L.~Sifre, G.~van~den Driessche, T.~Graepel, and D.~Hassabis.
\newblock {Mastering the Game of {Go} Without Human Knowledge}.
\newblock \emph{Nature}, 550:\penalty0 354--359, 2017.

\bibitem[van~den Herik and Winands(2008)]{vandenHerik2008}
H.~J. van~den Herik and M.~H.~M. Winands.
\newblock {Proof-Number Search and Its Variants}.
\newblock In \emph{Oppositional Concepts in Computational Intelligence}, pages 91--118. 2008.
\newblock ISBN 978-3-540-70829-2.

\bibitem[Winands and Bj\"{o}rnsson(2011)]{winands11}
M.~H.~M. Winands and Y.~Bj\"{o}rnsson.
\newblock {$\alpha\beta$-based Play-outs in {Monte-Carlo Tree Search}}.
\newblock In \emph{2011 IEEE Conference on Computational Intelligence and Games (CIG 2011)}, pages 110--117. IEEE, 2011.

\bibitem[Winands et~al.(2008)Winands, Bj{\"o}rnsson, and Saito]{winands08b}
M.~H.~M. Winands, Y.~Bj{\"o}rnsson, and J.-T. Saito.
\newblock {Monte-Carlo Tree Search Solver}.
\newblock In \emph{Computers and Games (CG 2008)}, volume 5131 of \emph{Lecture Notes in Computer Science (LNCS)}, pages 25--36, 2008.

\bibitem[Winands et~al.(2010)Winands, Bj\"{o}rnsson, and Saito]{winands10}
M.~H.~M. Winands, Y.~Bj\"{o}rnsson, and J.-T. Saito.
\newblock {Monte Carlo Tree Search in Lines of Action}.
\newblock \emph{IEEE Transactions on Computational Intelligence and AI in Games}, 2\penalty0 (4):\penalty0 239--250, 2010.

\bibitem[Świechowski et~al.(2023)Świechowski, Godlewski, Sawicki, and Mańdziuk]{swiechowski2023monte}
M.~Świechowski, K.~Godlewski, B.~Sawicki, and J.~Mańdziuk.
\newblock {Monte Carlo Tree Search}: {A} review of recent modifications and applications.
\newblock \emph{Artificial Intelligence Review}, 56\penalty0 (3):\penalty0 2497--2562, 2023.
\newblock \doi{10.1007/S10462-022-10228-Y}.

\end{thebibliography}

\end{document}